# Temporal Second Difference Traces

**Mitchell Keith Bloch**
University of Michigan
2260 Hayward Street
Ann Arbor, MI. 48109-2121
bazald@umich.edu

## Abstract

Q-learning is a reliable but inefficient off-policy temporal-difference method, backing up reward only one step at a time. Replacing traces, using a recency heuristic, are more efficient but less reliable. In this work, we introduce model-free, off-policy temporal difference methods that make better use of experience than Watkins' Q($\lambda$). We introduce both Optimistic Q($\lambda$) and the temporal second difference trace (TSDT). TSDT is particularly powerful in deterministic domains. TSDT uses neither recency nor frequency heuristics, storing $(s, a, r, s', \delta)$ so that off-policy updates can be performed after apparently suboptimal actions have been taken. There are additional advantages when using state abstraction, as in MAXQ. We demonstrate that TSDT does significantly better than both Q-learning and Watkins' Q($\lambda$) in a deterministic cliff-walking domain. Results in a noisy cliff-walking domain are less advantageous for TSDT, but demonstrate the efficacy of Optimistic Q($\lambda$), a replacing trace with some of the advantages of TSDT.

## 1 Introduction and Background

The focus of this work is on improving the efficiency of online, off-policy, temporal difference methods [Sutton and Precup, 1998; Watkins, 1989], without compromising stability of convergence. One-step methods such as Q-learning/Q(0) are slow but stable. Watkins' Q($\lambda$) is unstable for high decay rates and convergence is unproven in the general case.

We introduce two algorithms here. Optimistic Q($\lambda$) partially eliminates a disadvantage of Watkins' Q($\lambda$)–that the trace must be cleared when apparently suboptimal actions are taken. Temporal second difference traces (TSDT) fully eliminate this disadvantage of Watkins' Q($\lambda$) and are more stable when using state abstraction, as in MAXQ [Dietterich, 1998; Dietterich, 2000].

### 1.1 Basic Definitions

A state $s$ refers to a state in which an agent could find itself in the course of solving a problem. The state space $\mathcal{S}$ refers to the set of all states $s$. An action $a$ refers to a possible course of action for an agent. It can be useful to refer to a state-action pair, $(s, a)$, meaning to take action $a$ from state $s$. The action space $\mathcal{A}(s)$ refers to the set of all possible actions from state $s$. An absorbing state is one for which $\mathcal{A}(s)$ is empty. A terminal state is one in which the problem has been solved. A non-terminal state is one in which the process is incomplete, and therefore it is not an absorbing state. $\mathcal{S}^+$ refers to the subset of $\mathcal{S}$ containing only non-terminal states. $s'$ denotes a successor state following an action. A reward $r$ is a numerical value indicating the value of experiencing the triple $(s, a, s')$. A state transition refers to a state-action pair, a reward, and a successor state $(s')$, or the quadruple $(s, a, r, s')$.

A fully observable process is one in which an agent is always able to observe the $(s, a, r, s')$ quadruple. The state transitions for some problems can be modeled accurately with a probabilistic transition function, $\mathcal{P}^a_{ss'}$. The reward function, mapping $(s, a, s')$ to a reward, can sometimes be modeled accurately with a probabilistic reward function, $\mathcal{R}^a_{ss'}$. A Markov decision process (MDP) is a fully observable process in which $\mathcal{P}^a_{ss'}$ and $\mathcal{R}^a_{ss'}$ together provide an accurate model.

An online learning algorithm learns while gaining experience. An offline learning algorithm waits until it is finished gaining experience to learn. An on-policy learning algorithm learns about the policy it is currently following. An off-policy learning algorithm learns about a policy which may be different than that which it is currently following.

The discount rate $\gamma$ [0, 1] refers to how quickly an agent ceases to care about reward, and is often 1 for episodic/finite processes. Discounted return refers to the total discounted reward following a state-action pair. Expected return refers to the total discounted reward expected to follow a state-action pair. $\delta$ refers to the difference between expected discounted return and discounted return received. Temporal difference (TD) methods are online learning algorithms which update proportionally to $\delta$. The learning rate $\alpha$ (0, 1] refers to how much of the difference is applied in the update.

A state-action pair is starved if the state is never reached or the action is never attempted from the state. An exploration policy is non-starving if no state-action pair is starved as $t \to \infty$. A Q-value Q($s, a$) represents the current estimate for the expected return for a state-action pair. V($s$) refers to the maximum Q($s, a$) for all $a$. Q-learning/Q(0) is an off-policy TD method which updates one Q-value per step and is guaranteed to converge to the optimal policy given a non-starving exploration policy. Watkins' Q($\lambda$) is an off-policy eligibility trace which updates more than one Q-value per step [Watkins, 1989]. The decay rate $\lambda$ [0, 1] indicates how quickly entries in the trace cease to be updated as they become less recent.

## 1.2 Temporal Difference Methods

**On-Policy Backups**

$$Q(s,a) \Leftarrow Q(s,a) + \alpha[r + \gamma Q(s',a') - Q(s,a)] \quad (1)$$

Equation 1 is a standard one-step, on-policy TD backup. $Q(s,a)$ is updated to be closer to the sum of the immediate reward and the discounted return expected one step into the future. This can be expressed more readably:

$$Q(s,a) \xleftarrow{\alpha} r + \gamma Q(s',a') \quad (2)$$

This backup rule is used by Sarsa [Rummery and Niranjan, 1994], the canonical on-policy TD method to use Q-values. It describes the behavior of an agent navigating the state space using a somewhat greedy policy, and using equation 2 to learn from its experience. Being an on-policy algorithm, Sarsa learns about the actual policy being followed, incorporating the effects of exploration. Sarsa can be guaranteed to converge under certain conditions [Singh *et al.*, 1998].

**Off-Policy Backups**

There are strict requirements for Sarsa to converge to an optimal policy. Off-policy makes it easier to cope with more diverse exploration strategies. So long as $\alpha$ is sufficiently low and decreased appropriately for stochastic domains, the only requirement to guarantee convergence is that the exploration policy must be non-starving.

To accomplish this, Q-learning [Watkins, 1989] backs up the best next Q-value rather than the Q-value corresponding to the next selected action:

$$Q(s,a) \xleftarrow{\alpha} r + \gamma V(s') \quad (3)$$

Learning off-policy has the disadvantage that an agent may choose actions that are riskier given the exploration strategy, because the effect of exploration is completely removed. Another disadvantage is that techniques for speeding up learning become more difficult. Some of these difficulties will be discussed in section 1.3.

In exchange for these disadvantages, learning off-policy causes an agent's policy to more stably and directly approach the optimal policy, regardless of the exploration strategy. Furthermore, it enables an agent to learn about more than one policy at a time. This may not be a great advantage for flat reinforcement learning, but it can speed up hierarchical reinforcement learning considerably [Kaelbling, 1993].

**Terminal Backups**

Regardless of whether an agent is learning on-policy or off-policy, the expression is simpler still for a terminal backup:

$$Q(s,a) \xleftarrow{\alpha} r \quad (4)$$

This is automatic for problems for which all terminal states are absorbing states.

## 1.3 Beyond One-Step Methods

**Eligibility Traces**

Eligibility traces, such as Watkins' Q($\lambda$), are a model-free method for using recent memory to speed reinforcement learning. If one stores a trace of the state-action pairs taken over the course of a task, it is possible to pass $\delta$s back more than one step at a time. This can result in a significant increase in the speed of learning at a cost to stability.

Sarsa($\lambda$) [Rummery and Niranjan, 1994] is the standard on-policy eligibility trace. An entry can persist in the trace for arbitrarily many steps for $\gamma > 0$ and $\lambda > 0$, regardless of the rewards encountered.

Development of an off-policy eligibility trace is more difficult. When an agent takes an apparently suboptimal step for the sake of exploration, Q-values are updated on the basis of the Q-value of an action other than that which is taken.

Watkins' Q($\lambda$) [Watkins, 1989] is the standard off-policy eligibility trace. Entries are cleared from the trace after each apparently suboptimal action. Therefore, in the worst case, it is no more efficient at performing backups than Q(0). Entries can persist much longer in practice.

Peng's Q($\lambda$) [Peng and Williams, 1996] trades off some of the off-policy nature of Watkins's Q($\lambda$) in order to allow an entry to persist in the trace for arbitrarily many steps. Peng's Q($\lambda$) is neither on-policy nor off-policy.

**Dyna-Q**

As an alternative approach to speeding learning, Dyna-Q [Dyna, 1991] builds a model of the environment, learning both $\mathcal{P}^a_{ss'}$ and $\mathcal{R}^a_{ss'}$. Using this memory, it is able to learn from past experience. It can simulate either sample or full updates for arbitrary actions from visited states.

## 1.4 Paper Structure

Section 2 introduces Optimistic Q($\lambda$), an extension of Watkins' Q($\lambda$). Section 3 introduces the temporal second difference trace (TSDT), a different kind of memory trace with some of the properties of eligibility traces. Section 4 presents experimental results for both algorithms in two cliff-walking domains. Section 5 provides a discussion of theory and results.

# 2 Optimistic Q($\lambda$)

Optimistic Q($\lambda$) alleviates the need to completely clear traces as in Watkins's Q($\lambda$). However, it allows only positive net updates to take place after apparently suboptimal actions have been taken.

Optimistic Q($\lambda$) as depicted in algorithm 1 is the first of two traces developed in this paper. It is based on Watkins' Q($\lambda$). The algorithm is extended to track return experienced past an apparently suboptimal action. If the sum of the return experienced so far and the expected best return for the actions currently available would increase a Q-value, then the Q-value is updated even if an apparently suboptimal action has been taken since the entry was added to the trace. This is sound because the update is performed only if the apparently suboptimal choice of action ends up appearing optimal given information gained later on.

**Algorithm 1** Optimistic Q($\lambda$). $O(s,a)$ stores whether a given Q-value must be updated optimistically only. $E(s,a)$ stores the partial return experienced since $O(s,a)$ became $True$.

**Ensure:** $Q$ initialized arbitrarily, e.g., $Q(s,a) = 0$,
  for $\forall s \in \mathcal{S}^+, \forall a \in \mathcal{A}(s)$
1: **while** an episode is to occur **do**
2:   Initialize $s$ {non-terminal, non-starving}
3:   Initialize $e(s,a) = 0$ for $\forall s \in \mathcal{S}^+, \forall a \in \mathcal{A}(s)$
4:   Choose $a$ from $\mathcal{A}(s)$ {non-starving}
5:   **repeat** {for each step of the episode}
6:     **if** $Q(s,a) < V(s)$ **then**
7:       **for all** $s \in \mathcal{S}, a \in \mathcal{A}(s)$ **do**
8:         $O(s,a) \Leftarrow True$ {Instead of $e(s,a) \Leftarrow 0$}
9:       **end for**
10:    **end if**
11:    Take action $a$, observe reward, $r$, and next state, $s'$
12:    **for all** $b \in \mathcal{A}(s)$ **do** {Replacing trace}
13:      **if** $b = a$ **then**
14:        $e(s,b) \Leftarrow 1$
15:        $O(s,b) \Leftarrow False$
16:      **else**
17:        $e(s,b) \Leftarrow 0$
18:      **end if**
19:    **end for**
20:    Choose $a'$ from $\mathcal{A}(s')$ {non-starving}
21:    $\delta_{\text{on}} \Leftarrow \gamma Q(s', a') - Q(s,a)$
22:    $\delta_{\text{off}} \Leftarrow \gamma V(s') - Q(s,a)$
23:    **for all** $s \in \mathcal{S}, a \in \mathcal{A}(s)$ **do**
24:      **if** $O(s,a) = False$ **then**
25:        $E(s,a) \Leftarrow 0$
26:      **end if**
27:      $E(s,a) \Leftarrow E(s,a) + e(s,a)r$
28:      $\delta \Leftarrow E(s,a) + e(s,a)\delta_{\text{off}}$
29:      **if** $O(s,a) = False$ **or** $\delta > 0$ **then**
30:        $Q(s,a) \xleftarrow{\alpha} Q(s,a) + \delta$
31:        $E(s,a) \Leftarrow e(s,a)(\delta_{\text{on}} - \delta_{\text{off}})$
32:        {Optionally $O(s,a) \Leftarrow False$}
33:      **end if**
34:      $e(s,a) \Leftarrow \gamma \lambda e(s,a)$
35:    **end for**
36:    $s \Leftarrow s'$ and $a \Leftarrow a'$
37:  **until** $s$ is terminal
38: **end while**

**Algorithm 2** Temporal Second Difference Trace (TSDT). Note that $\delta^2$ is the second difference. $\delta^2 \neq \delta \cdot \delta$.

**Ensure:** $Q$ initialized arbitrarily, e.g., $Q(s,a) = 0$,
  for $\forall s \in \mathcal{S}^+, \forall a \in \mathcal{A}(s)$
1: **while** an episode is to occur **do**
2:   Initialize $s$ {non-terminal, non-starving}
3:   Initialize $t(s,a) = \emptyset$ for $\forall s \in \mathcal{S}^+, \forall a \in \mathcal{A}(s)$
4:   **repeat** {for each step of the episode}
5:     Choose $a$ from $\mathcal{A}(s)$ {non-starving}
6:     Take action $a$, observe reward, $r$, and next state, $s'$
7:     **for all** $b \in \mathcal{A}(s)$ **do** {Replacing trace}
8:       **if** $b = a$ **then**
9:         $t(s,b) \Leftarrow t$
10:        $r(s,b) \Leftarrow r$
11:        $s'(s,b) \Leftarrow s'$
12:        $\delta(s,b) \Leftarrow 0$
13:      **else**
14:        $t(s,b) \Leftarrow \emptyset$
15:      **end if**
16:    **end for**
17:    **for all** $s \in \mathcal{S}, a \in \mathcal{A}(s), t(s,a_i) \neq \emptyset$,
         in reverse $t$ order **do**
18:      $\delta \Leftarrow r(s,a) + \gamma V(s'(s,a)) - Q(s,a)$
19:      $\delta^2 \Leftarrow \delta - \delta(s,a)$
20:      $Q(s,a) \Leftarrow Q(s,a) + \alpha \delta^2$
21:      $\delta(s,a) \Leftarrow r(s,a) + \gamma V(s'(s,a)) - Q(s,a)$
22:    **end for**
23:    $s \Leftarrow s'$
24:  **until** $s$ is terminal
25: **end while**

## 3 Temporal Second Difference Trace

Eligibility traces and Dyna-Q are well known mechanisms for speeding up reinforcement learning. Unfortunately, attempts to apply eligibility traces to off-policy learning have been limited in their success. Eligibility traces have been cut short [Watkins, 1989], given up on being entirely off-policy [Peng and Williams, 1996], and become very complicated [Precup *et al.*, 2000]. Dyna-Q is both simple and powerful but requires the agent to learn a model ($\mathcal{P}_{ss'}^a$ and $\mathcal{R}_{ss'}^a$). Here we introduce an algorithm with none of these limitations.

### 3.1 Not (Quite) an Eligibility Trace

The temporal second difference trace (TSDT), described in algorithm 2, is our version of a memory trace. It is interesting in that it isn't an eligibility trace in the usual sense. Rather than keeping track of the eligibility or strength of entries in the trace, the temporal second difference trace simply keeps track of the updates being performed. Using this information, it is able to tweak the updates as information becomes available. Intuitively, it is as though earlier updates, based on less complete information, are redone using more complete information. It does not need to track an eligibility (or strength) value at all.

More specifically, line 18 calculates the difference between the current Q-value and what it is tending towards. Line 19 calculates the (second) difference between this difference and

Let us step through the key part of the algorithm. Line 27 accumulates the return experienced since the Q-value was last updated. Line 28 adds in the off-policy (best) value for the current state. Line 29 allows the update to take place only if no apparently suboptimal actions have been taken since the Q-value was added to the trace, or if the update is positive enough to be better than the last update, including the off-policy update. Line 30 does the straightforward step of updating the Q-value. Line 31, however, stores the negative off-policy part of the update, causing the math in lines 29 and 30 to work out in the case that updates must be optimistic. In the case that updates need not be optimistic, line 25 later resets the value.

the previous difference. Line 20 updates the Q-value proportionally to the second difference. Line 21 then stores the new difference. $\delta^2$ is non-zero if either $Q(s,a)$ or $V(s')$ has shifted, though the former can only happen when duplicate Q-values appear in the trace.

### 3.2 Efficient Propagation of Information

A parameter, $t$, is used to facilitate updating the trace in reverse chronological order. As in Watkins' $Q(\lambda)$ and Dyna-Q this is not strictly necessary, but it guarantees that information will propagate backwards though the trace as efficiently as possible. At each step, the trace simply recalculates the difference, $\delta$, and then modifies the Q-value by the second difference, $\delta^2$. Updates are as direct as those performed by Dyna-Q, but no model is necessary. Regardless, a model could be used to eliminate the learning rate parameter.

For the sake of computational efficiency, one can limit the length of the trace or speed the implementation using lazy updates and backward replay as with Watkins' $Q(\lambda)$.

### 3.3 Some Examples

Figure 1 depicts an MDP where state C is the focus. Let us examine the results of learning with Watkins' $Q(\lambda)$, Optimistic $Q(\lambda)$, and the Temporal Second Difference Trace (TSDT) across several episodes. We use the syntax $\{s_0,s_1,\ldots,\text{terminal reward}\}$ for an episode. We use the rates $\alpha = 1$, $\gamma = 1$, and $\lambda = 1$ where applicable.

**Advantage of Longer Traces**

Having a longer trace gives Optimistic $Q(\lambda)$ and TSDT the ability to learn from rewards further in the future than can Watkins' $Q(\lambda)$ when learning off-policy.

Let us examine what happens when episode $\{A,C,1\}$ is followed by episode $\{A,C,10\}$. AC results in $\delta = -1$, causing $Q(A,C) = -1$. C-1 results in $\delta = 1$, causing $Q(C,1) = 1$ and $Q(A,C) = 0$. The trace is cleared between episodes. AC results in $\delta = 0$. Finally C-10 results in different outcomes for the algorithms. Watkins' $Q(\lambda)$ clears the trace, Optimistic $Q(\lambda)$ sets all entries to update optimistically only, and TSDT simply keeps all entries in the trace. Given $\delta = 10$, $Q(C,10) = 10$ for all algorithms. Having cleared its trace, Watkins' $Q(\lambda)$ is unable to update $Q(A,C) = 9$, but Optimistic $Q(\lambda)$ is able to perform the update, given that $10 - 1 > 0$. TSDT also performs the update given that $V(C)$ has increased.

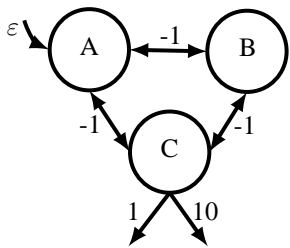

Figure 1: A deterministic MDP in which an agent begins in state A, moves freely between states A, B, and C, and finally terminates in state C.

**Advantage of the Second Difference**

That TSDT does updates based on a one-step backup rule gives it an advantage over Watkins' $Q(\lambda)$ and Optimistic $Q(\lambda)$ which rely instead on a recency heuristic [Singh *et al.*, 1996].

Let us examine what happens when episode $\{A,C,1\}$ is followed by episode $\{A,C,B,C,10\}$. Let us jump ahead to CB. Watkins' $Q(\lambda)$ clears the trace, Optimistic $Q(\lambda)$ sets all entries to update optimistically only, and TSDT simply keeps all entries in the trace. Given $\delta = -1$, $Q(C,B) = -1$ for all three algorithms. BC results in $\delta = -1$. $Q(B,C) = -1$ for all three algorithms. $Q(C,B) = -2$ for both eligibility traces, but remains unchanged for TSDT given that $Q(BA)$ is still 0. Finally C-10 results in different outcomes for all three algorithms. Once again Watkins' $Q(\lambda)$ clears the trace, Optimistic $Q(\lambda)$ sets all entries to update optimistically only, and TSDT simply keeps all entries in the trace. Given $\delta = 10$, $Q(C,10) = 10$ for all algorithms. Watkins' $Q(\lambda)$ is unable to update $Q(B,C)$, $Q(C,B)$, or $Q(A,C)$. Optimistic $Q(\lambda)$ updates $Q(B,C) = 9$, $Q(C,B) = 8$, and $Q(A,C) = 7$. TSDT updates $Q(B,C) = 9$, $Q(C,B) = 8$, and $Q(A,C) = 9$. Note that here Optimistic $Q(\lambda)$ comes closer to converging than does Watkins' $Q(\lambda)$, and that TSDT does even better.

**Gracefully Handling State Abstraction**

Watkins' $Q(\lambda)$ and Optimistic $Q(\lambda)$ both pass back the $\delta$ for the current $Q(s,a)$. Each entry in the trace gets updated in proportion to $\alpha\gamma e(s,a)$, $e(s,a)$ being a function of lifetime in the trace. Given that $\alpha$, $\gamma$, and $\lambda$ can all be 1, $\alpha\gamma e(s,a)$ can be 1 for the lifetime of the entry in the trace. Thus, if part of a Q-value applies to more than one state due to state abstraction, an entry in the trace can experience the $\delta$ arbitrarily many times. A small change in expectation for the action just taken can be magnified many fold. In TSDT, $\delta$ is not passed back at all. Rather, a local $\delta$ is calculated for each entry of the trace. If the $\delta$ has changed since the last time the entry was updated, the Q-value is updated proportionally to the change in $\delta$. In a case that would cause an update to be magnified in a regular eligibility trace, both $Q(s,a)$ and $V(s'(s,a))$ will shift in at least one entry of the trace. This will cause the second difference to safely approach zero.

Let us discuss what happens when episode $\{A,C,1\}$ is followed by episode $\{A,C,B,C,10\}$ if a value is shared between $Q(A,C)$ and $Q(B,C)$. Let us jump ahead to BC. This time nothing happens given $Q(B,C) = -1$ already and $\delta = 0$. C-10 however causes Optimistic $Q(\lambda)$ to double count $\delta = 10$, first updating $Q(B,C) = 9$ and then $Q(A,C) = 16$. Watkins' $Q(\lambda)$ avoided this problem by clearing the trace earlier, and similarly the problem can be avoided here by evicting duplicate Q-values from the trace rather than relying purely on the usual replacing trace semantics. However, TSDT does not suffer from this problem at all, updating $Q(B,C) = 9$ and then leaving the value unchanged when updating $Q(A,C) = 9$.

This is not to say that TSDT completely solves all problems resulting from having duplicate Q-values in a trace. When using $\alpha < 1$, having duplicate entries can either decrease or increase the effective learning rate for the Q-value. Having different rewards or transitions for a Q-value in a trace can result in an effective decrease in $\alpha$. Having duplicate rewards and transitions for a Q-value in a trace too near to one another can result in an effective increase in $\alpha$. For this reason, it remains important to eliminate duplicate entries or bound TSDT sizes when using $\alpha < 1$, as with eligibility traces, though the problem is less severe for TSDT. These problems are entirely absent when using $\alpha = 1$ for deterministic processes.

## 4 Experimental Results

Experimental results presented are an average of 30 different sets of episodes, each starting with a different random seed. Further, each plot is smoothed with a running average of 200 episodes.

Given that these problems are tractable using value iteration, the graphs plot the total suboptimality of all actions for a given episode. In other words, they plot $\sum_t [Q(s_t, a_t) - V(s_t)]$.

As both versions of the domain examined are episodic, a discount rate of 1 is used.

### 4.1 Deterministic Cliff-Walking Domain

The cliff-walking domain provides a useful testbed because it provides many opportunities for failure, some of which are very close to the goal. Additionally, it seems intuitive that an agent with a maximally effective memory trace should be able to learn a good path from any particular state in one episode, should it happen to cover the right ground.

In the domain depicted in figure 2 there are 49 non-terminal states, 1 goal state (marked with an 'X'), and 9 failure states. Four actions are allowed from each non-terminal state, each of which deterministically moves the agent 1 tile in the specified direction if possible. Arriving at the goal yields 20 reward, walking off the cliff yields $-20$ reward, and all other transitions yield $-1$ reward.

All four agents tested in this domain use a fixed epsilon-greedy exploration strategy, with $\varepsilon = 0.3$, preventing any of the agents from behaving optimally during exploration. Q-learning, an established off-policy algorithm guaranteed to converge on an optimal policy, nearly finishes converging after 1500 episodes. The temporal second difference trace (TSDT) nearly finishes converging in in only 1000 episodes, and without the early dip in performance experienced by Watkins' Q(1). Both eligibility trace methods, however, result in divergent behavior for hundreds of thousands of episodes.

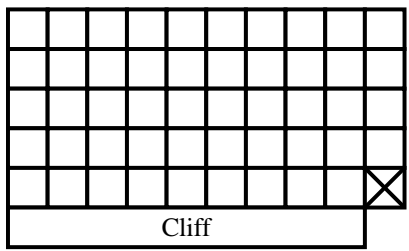

Figure 2: The cliff-walking domain explored here, with and without noise affecting move actions.

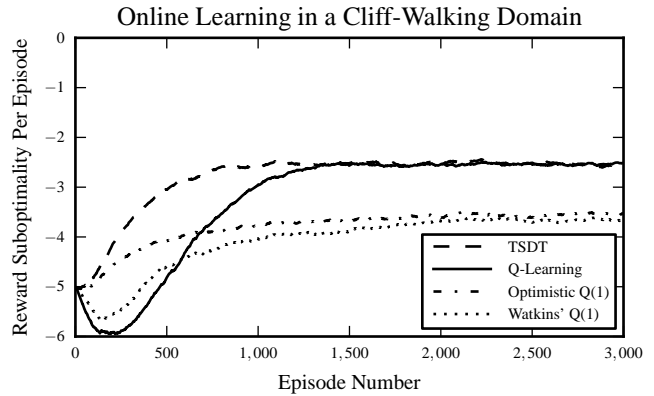

Figure 3: This plot depicts the online performance of agents following a fixed epsilon-greedy exploration strategy in a cliff-walking domain.

While this plot depicts a running average (over 200 episodes) of 30 different sets of 2000 episodes for each temporal difference method, more detailed statistics are of some interest. Across all 30 policies developed for each of the 49 different initial conditions, Q-learning and TSDT have optimal policies for all instances. Watkins' Q(1) and Optimistic Q(1) have optimal policies for only 360 and 289 respectively of the 1470 instances, however.

Results not reproduced here indicate that $\lambda \approx 0.2$ allows the eligibility trace methods to stably converge on optimal policies, although not much faster than Q-learning.

### 4.2 Noisy Cliff-Walking Domain

We now introduce a version of the domain in which actions may result in different state transitions (and the corresponding rewards) with some probability. For this experiment, the transition will behave normally with probability $0.8$ or its direction will be rotated $90°$ clockwise or counter-clockwise with equal probabilities $0.1$.

Strictly speaking, carefully decreasing the learning rate is required for convergence in a stochastic MDP. However, we will use a fixed $\alpha = 0.05$ and the same exploration strategy as in the previous experiment, acknowledging that unlearning is possible.

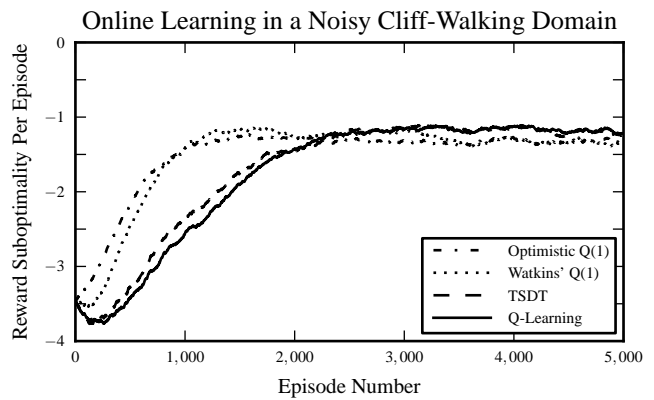

Figure 4: This plot depicts the online performance of agents following a decreasing epsilon-greedy exploration strategy in a noisy cliff-walking domain.

Here Optimistic Q(1) and Watkins' Q(1) initially do very well, compensating for the low learning rates. However, TSDT and Q-learning eventually overtake both eligibility traces. At the end of 5,000 episodes, Q-learning has the best policy, Optimistic Q(1) has finally caught up with TSDT again, and Watkins' Q(1) has the least optimal policy. Due to the decreased learning rate, all methods are more even in the stochastic domain, but both Q-learning and TSDT approach equilibrium more directly than the eligibility trace methods.

## 5 Discussion

Temporal second difference traces (TSDT) have been demonstrated to be immune to some of the flaws of Watkins' Q($\lambda$). TSDT has no need to clear traces to ensure the validity of off-policy learning when apparently suboptimal actions are taken. However, care must still be taken to bound the number of duplicate Q-values when learning stochastic processes.

Optimistic Q($\lambda$) is a replacing eligibility trace with some of the advantages of TSDT. Traces do not need to be cleared after apparently suboptimal actions, and increasing expectations of reward can overcome the penalty for this exploration. However, decreasing expectations of reward cannot generally be passed back through the trace beyond apparently suboptimal actions as in TSDT. Additionally, the advantages of TSDT with respect to state abstraction are lost. Despite these limitations, Optimistic Q($\lambda$) seems to be effective when learning stochastic processes.

TSDT has been demonstrated to converge on the optimal policy for the deterministic cliff-walking domain significantly faster than Q-learning, as opposed to Q(1) which exhibits significant divergent behavior. TSDT has been shown to be more comparable to Q-learning than Q(1) in a noisy cliff-walking domain as well, though the Q(1) methods do better early on. Additionally, Optimistic Q(1) outperformed Watkins' Q(1) slightly in both cliff-walking domains.

We expect TSDT to perform as least as well as Q($\lambda$) for deterministic domains. However, it may not make as much use of information as Q($\lambda$) when learning rates are low and the traces are long. The amount of return used by TSDT decreases exponentially with respect to the learning rate, as opposed to Q($\lambda$) which decreases exponentially only if $\lambda < 1$. Therefore it is important for the efficiency of TSDT to use higher learning rates whenever possible. It may be that decreasing the learning rate per Q-value with respect to $1/n$ could be sufficient to significantly improve the efficiency of TSDT in stochastic domains, though alternatives which may keep $\alpha$ higher longer could do better still.

We believe that using TSDT instead of Q-learning and Optimistic Q($\lambda$) instead of Watkins' Q($\lambda$) should be reasonable regardless of the domain. The only downside is increased computational cost.

We have done additional research in the area of hierarchical reinforcement learning, looking at the taxicab domain [Dietterich, 1998] and the fickle taxicab domain [Dietterich, 2000]. Our research has continued to focus on off-policy learning. We expect to present this additional work using TSDT in a future publication.

## Acknowledgments

I would like to thank Professor John Laird and the University of Michigan for their support. I would like to thank the Soar Group, including Professor John Laird, Jon Voigt, Nate Derbinsky, Nick Gorski, Justin Li, Bob Marinier, Shiwali Mohan, Miller Tinkerhess, Yongjia Wang, Sam Wintermute, Joseph Xu, and Mark Yong for helping me to refine the presentation of these ideas. Additionally, I would like to thank Professor Satinder Singh for meeting with me to discuss some of his past work.